\documentclass[lettersize,journal]{IEEEtran}
\usepackage{amsmath,amsfonts}
\usepackage{algorithmic}
\usepackage{algorithm}
\usepackage{array}
\usepackage[caption=false,font=normalsize,labelfont=sf,textfont=sf]{subfig}
\usepackage{color}
\usepackage{textcomp}
\usepackage{stfloats}
\usepackage{url}
\usepackage{verbatim}
\usepackage{graphicx}
\usepackage{cite}
\usepackage[numbers]{natbib}
\definecolor{aqua}{rgb}{0.0, 1.0, 1.0}

\begin{document}

\title{MultiScale Probability Map guided Index Pooling with Attention-based learning for Road and Building Segmentation}

\author{Shirsha~Bose, Ritesh~Sur~Chowdhury, Debabrata~Pal\\
Shivashish~Bose,
Biplab~Banerjee,~\IEEEmembership{Member,~IEEE},
and~Subhasis~Chaudhuri,~\IEEEmembership{Fellow,~IEEE}

\thanks{Shirsha Bose is with Department of Informatics, Technical University of Munich, Germany (email: shirshabosecs@gmail.com). Ritesh Sur Chowdhury is with Department of Electronics and Telecommunication Engineering, Jadavpur University, India (email: riteshsurchowdhury2001@gmail.com). Debabrata Pal is associated with IIT Bombay, India (email: deba.iitbcsre19@gmail.com). Biplab Banerjee is Associate Professor at the Center for Studies in Resources Engineering (CSRE) and CMINDS, IIT Bombay, India (email: getbiplab@gmail.com). Shivashish Bose is Professor at the Architecture Department of Jadavpur University, India (email: shivashish.bose@jadavpuruniversity.in). Subhasis Chaudhuri is Professor at the Dept of EE, IIT Bombay, India (email: sc@ee.iitb.ac.in).

Shirsha Bose and Ritesh Sur Chowdhury contributed equally to this work. 
}
}



\maketitle

\begin{abstract}
Efficient road and building footprint extraction from satellite images are predominant in many remote sensing applications.
However, precise segmentation map extraction is quite challenging due to the diverse building structures camouflaged by trees, similar spectral responses between the roads and buildings, and occlusions by heterogeneous traffic over the roads. Existing convolutional neural network (CNN)-based methods focus on either enriched spatial semantics learning for the building extraction or the fine-grained road topology extraction. The profound semantic information loss due to the traditional pooling mechanisms in CNN generates fragmented and disconnected road maps and poorly segmented boundaries for the densely spaced small buildings in complex surroundings. In this paper, we propose a novel attention-aware segmentation framework, Multi-Scale Supervised Dilated Multiple-Path Attention Network (MSSDMPA-Net), equipped with two new modules Dynamic Attention Map Guided Index Pooling (DAMIP) and Dynamic Attention Map Guided Spatial and Channel Attention (DAMSCA) to precisely extract the building footprints and road maps from remotely sensed images. DAMIP mines the salient features by employing a novel index pooling mechanism to retain important geometric information.
On the other hand, DAMSCA simultaneously extracts the multi-scale spatial and spectral features. Besides, using dilated convolution and multi-scale deep supervision in optimizing MSSDMPA-Net helps achieve stellar performance. Experimental results over the seven benchmark building and road extraction datasets namely, Porto, Shanghai, Massachusetts Road, Massachusetts Building, Synthinel-1, WHU Satellite I and WHU Ariel Imagery dataset, ensures MSSDMPA-Net as the state-of-the-art (SOTA) method for building and road extraction as our method beats the next best alternatives by 5.94\%, 2.55\%, 3.97\%, 11.64\%, 6.86\%, 6.92\%, 2.57\% IOU and 3.98\%, 1.90\%, 2.43\%, 7.17\%, 3.98\%, 4.99\%, 1.37\%  F1 score, respectively, over the mentioned datasets.                  
\end{abstract}


\section{Introduction}
Due to fast-paced urbanization, man-made roads, and buildings are ever-evolving in our society. They have become the top-most concerns to monitor for topographic mappings continuously, change detection, emergency services for disaster management \citep{adriano2021learning, harb2017remote}, smart-city development, autonomous driving, 3-D reconstruction of the terrain, etc. The high-resolution optical satellite images periodically provide granular mapping with sharp convex boundaries of the buildings. The geoscience community has widely adapted the road networks' well-connectivity along with the accurate building footprint and road network extraction, as the world is driving toward autonomous solutions. In this endeavor, automatic map extraction holds much more prominence than conventional manual road labeling, GPS trajectories aggregation, or map extraction from LIDAR point clouds \citep{mattyus2017deeproadmapper, etten2020city}. However, to fully complement the benefits of high-resolution satellite images for automatic map extraction, it also requires simultaneous advancement of the segmentation methods \citep{ronneberger2015u, wei2019toward, zhu2020map}.

Conventional handcrafted building and road feature extraction methods \citep{zhou2014seamless, stoica2004gibbs, laptev2000automatic} heavily depend on the spectral responses, texture, geometry, and shadow characteristics. However, these methods exhibit generalization inefficacy in recognizing complex polygon-shaped buildings with varying scales and roads with spectrally diversified materials in occlusion, illumination, and sensor variations. The recent advancements in CNN-based map extraction \citep{mei2021coanet, chaurasia2017linknet, wei2019toward} created a profound interest in the geoscience community due to the power of generalization. Initially, the fully-convolutional network (FCN)-based road segmentation \citep{cheng2017automatic, li2018road, tsutsui2017distantly} method suffered from road connectivity fragmentation. U-Net-based architecture \citep{lu2021gamsnet, chaurasia2017linknet, mosinska2018beyond} and iterative refinement in post-processing \citep{mnih2012learning, wegner2015road} tried to address the road map fragmentation problem by judiciously exploiting a larger image context while labeling the pixels. 
Nonetheless, it is worth mentioning that the existing research disjointly learns segmenting either the road networks or the buildings. Specifically, methodologies for road network extraction learn the graph connectivity of various road topologies and often fail to extract building boundaries or detect small buildings accurately. Similarly, building segmentation methodologies produce missing connections while predicting road networks. \textit{As a consequence, no unified methodology exists that can equally segment the road connectivity along with the sparsely spaced tiny buildings to the best of our knowledge.}

A critical aspect of the existing CNN-based segmentation models is considering pooling layers within the network using generally, max-pooling or average pooling, to down-sample a feature map's spatial dimension. However, this leads to heavy semantic information loss, specifically for dense prediction tasks, including segmentation. Mathematically, max-pooling only retains the high-intensity feature, whereas average pooling smooths out the features by computing the mean operation. Likewise, Mixed Pooling \citep{yu2014mixed} and Hybrid Pooling \citep{lee2016generalizing} strategies are deemed to combine max pooling and average pooling to perform feature down-sampling. Few other pooling mechanisms, stochastic pooling \citep{zeiler2013stochastic}, and pyramid pooling \citep{zhu2020map, he2019road, peng2020semantic} have also been explored for precise segmented map generation. Recently, a unified framework \citep{gao2019lip} aggregates features with their local importance in each stride of the sliding window to preserve local information. Finally, it is possible to use the multi-strided convolution operation for spatial down-sampling feature maps \citep{he2016deep}. Despite the advancements, the pooling layers are found to inversely affect the extraction of complex objects like buildings and roads by producing fragmented road segments or over-smoothed building boundaries.
This leads us to ask the research question: \textit{how to reduce the semantic information loss in the pooling operation in CNN?}

\begin{figure}
\centering
\includegraphics[height=5cm, width=7cm]{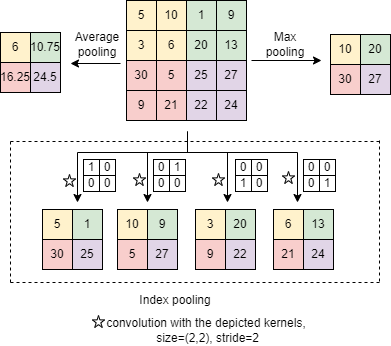}
\vspace*{-4mm}
\caption{Illustration of the different pooling mechanisms: Average pooling, Max pooling and Index pooling. Convolution of a feature map with different binary kernels produces multiple downsampled images. These kernels contain a value of `one' at a unique index position, and the remaining are filled with 'zero'. Then, the downsampled features are concatenated channel-wise. The index pooling layer preserves all the feature information and eliminates the semantic information loss while performing the spatial dimension reduction. For ease of understanding, we have shown four kernels of size (2, 2) and stride value of 2 to downsample the feature map by selecting one unique indexed value in a 2 x 2 window.}
\vspace*{-2mm}
\label{fig1}
\end{figure}

As a remedy, we propose a novel segmentation architecture MSSDMPA-Net, capable of segmenting building footprints and road networks from high-resolution satellite images. Furthermore, to tackle the conventional issues in pooling the intermediate feature maps from our CNN-based model, we propose a novel attention-guided pooling operation named Dynamic Attention Map guided Index Pooling (DAMIP). The DAMIP module consists of the novel lossless Index Pooling operation, which is lossless as the exact feature information is distributed across spatially downsampled multiple features as shown in Fig. \ref{fig1}. Moreover, these generated feature maps are then highlighted by the attention mechanism of the DAMIP leveraging our network's prior generated segmentation maps towards the generation of semantic information-preserving context-aware feature maps of reduced spatial resolution.


From another perspective, it is a fact that attention-based learning improves salient feature extraction while suppressing irrelevant features \cite{park2018bam, woo2018cbam, hu2018squeeze}. Attention mechanisms have also been infused with segmentation algorithms in \citep{zhu2020map,li2021multiattention,gao2021road,cai2021mha}, to extract better road and building segmentations, mostly by averaging or maximizing salient features over the channel dimensions of the feature maps. For example, CoANet \citep{mei2021coanet} applies a connectivity attention module based on Squeeze-and-excitation networks (SE-Net) \citep{hu2018squeeze}. 
In \citep{gao2021road, li2021multiattention}, spatial and channel attention modules are utilized based on the scaled-dot product attention of \citep{zhang2019self}. However, despite its usage, it is seldom found that attention learning highlights some domain-dependent information like image backgrounds.
If the attention module focuses on irrelevant parts of the feature map, this error gets propagated throughout the network, resulting in poor segmented map generation. Motivated by this research gap, we ask: \textit{how to learn a semantically meaningful and discriminative attention map for the segmentation task?}

To suppress irrelevant information broadcasting and preserve the geometric structure of the urban objects, we generate prior segmented probability maps through multi-scale supervision. Then, these probability maps provide salient region information to the feature maps via the Dynamic Dynamic Attention Map Guided Spatial and Channel Attention (DAMIP) module. The DAMIP consists of spatial and channel attention modules whose attention mask is the generated segmentation maps. This makes the mutli-scale features spatially and spectrally aware of the exact semantic context of the feature maps required for accurate road network and building footprint extraction. Furthermore, the generated segmentation maps also provide semantic awareness to the lossless index-pooled feature maps at the DAMIP module.


By design, MSSDMPA-Net follows a multi-path network architecture, which is influenced by the working principles of HRNet \cite{wang2020deep}. By design, HRNet fuses cross-stream convolutional multi-resolution features in parallel, high to low, and low to high-level features to generate high-resolution segmentation maps. But this increases the parameters and the overall complexity, making it prone to vanishing gradients. However, instead of the cross-stream convolutional operations, we process feature maps of different spatial resolutions in each path of our multipath framework. In each path, multiple dilated residual convolutional units with increasing receptive fields help preserve the granular geometrical characteristics of the remotely sensed buildings and road networks from widely varied spatial resolutions. Then the output feature maps of each multipath encoder are supervised by the novel Dynamic Probability Map Generator (DPMG) module to generate the segmented probability maps of various resolutions and provide deep supervision to the whole multipath encoder. By leveraging the DPMG module-generated segmented probability maps, DAMIP performs attention-based learning with the semantic information-preserved down-sampled feature maps. We subsequently up-sample multi-scale features to a higher resolution using the DAMSCA module utilizing the segmented probability maps again. Finally, a decoder generates the final segmented map from concatenated DAMSCA features. We summarize our significant contributions as follows,

\noindent - We introduce a novel multi-path attention-aware network, MSSDMPA-Net, to extract the salient hierarchical structures of the urban objects from remote sensing scenes via incremental dilated convolutions. The rigorous supervision of the multi-scale feature learning framework within MSSDMPA-Net helps to alleviate the vanishing gradient problem and makes the network semantically aware toward precise geometric structures.

\noindent - We propose a novel multi-scale supervised probability maps guided index-pooling mechanism based on attention learning to downsample a feature map spatially without semantic information loss. Also, leveraging the multi-scale supervised probability maps, our novel DAMIP module preserves intact semantic information and contextual awareness.

\noindent - Our novel spatio-spectral attention module, DAMSCA, utilizes the multi-scale supervised probability maps to produce geometry-aware upsampled feature maps.

\noindent - In addition to showing that the proposed MSSDMPA-Net outperforms the existing state-of-the-art methods after evaluating over seven benchmark datasets on the road and building segmentation, we perform rigorous ablation analysis.

\begin{figure*}[!htbp]
\includegraphics[width=\linewidth]{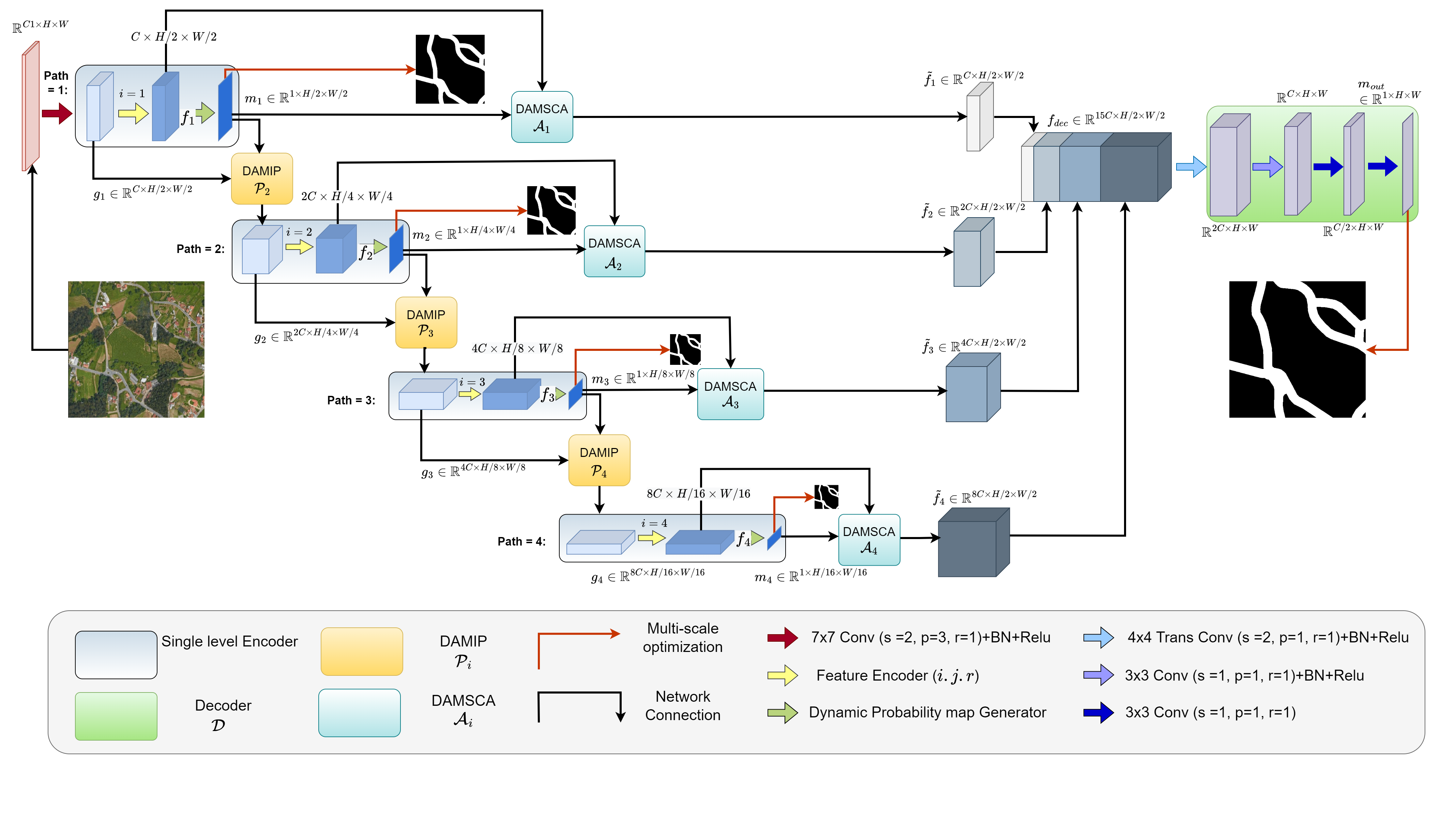} 
\vspace*{-15mm}
\caption{The architecture of the proposed MSSDMPA-Net contains four major modules, namely, the multi-path encoder (${\mathcal{H}_i}$), DAMIP (${\mathcal{P}_i}$), DAMSCA (${\mathcal{A}_i}$), and decoder (${\mathcal{D}}$). Each of the four paths is denoted by $i$ where $i\epsilon\{1,2,3,4\}$. $j$ is the number of dilated convolution blocks where $j\epsilon\{1,2,3,4\}$. Each path of the four multi-path encoders comprises a feature encoder ($\mathcal{G}_i$) having several dilated convolution blocks connected in series with incremental dilation rate and a DPMG block ($\mathcal{F}_i$). Again, each of the three DAMIP modules ($\mathcal{P}_i$) at individual paths performs attention-based learning and downsamples the spatial resolution of the feature map. On the other hand, DAMSCA modules ($\mathcal{A}_i$) also perform attention-based learning but upsample the feature maps. Finally, the DAMSCA modules generated features are concatenated ($f_{dec}$) and passed to the decoder ($\mathcal{D}$) to generate a segmented probability map. Using multi-scale supervision, we jointly optimize the multi-path encoders generated features ($m_{i}$) and the decoder-produced probability map ($m_{out}$).}
\vspace*{-2mm}
\label{fig2}
\end{figure*}

\section{Related works}

\noindent \textbf{Road and Building segmentation:} The RS literature is rich in CNN-based road extraction frameworks. In CoANet \citep{mei2021coanet}, the authors have developed an encoder-decoder-based road extraction framework where a Strip Convolution Module (SCM) is used to learn the long-range dependencies in road regions from four different directions. To improve accuracy and limit the model weight, D-Linknet \citep{zhou2018d} followed Linknet \citep{chaurasia2017linknet} by making a direct residual connection from the encoder to the decoder and used dilated convolutions to increase the receptive field. DAD-LinkNet \citep{gao2021road} adaptively integrated the local and global road features by using floating vehicle trajectory and satellite data jointly. In \citep{shamsolmoali2020road}, the authors incorporated a feature pyramid network in the generative adversarial networks to minimize the difference between the source and target domains for road segmentation. A considerable performance from encoder-decoder-based framework are observed in \citep{mei2021coanet,gao2021road,zhou2018d} for road segmentation.

Similar to the road segmentation networks, several building segmentation endeavors \citep{kang2019eu, shao2020brrnet, chen2022res2} also leveraged the U-Net framework for precisely segmenting the building footprints. In \citep{wei2019toward}, the authors proposed a scale robust fully convolutional network (FCN) equipped with a polygon regularization algorithm. In \citep{ji2018fully}, the authors introduced a Siamese U-Net with two parallel inputs, one original image, and a downsampled counterpart, thereby optimizing the network with shared weights. In RAPNet \citep{tian2021multiscale}, CBAM attention refines pyramid pooled feature maps, followed by a dense connection to embed high-level information into low-level information.

MAP-Net \citep{zhu2020map} and MANet \citep{li2021multiattention} have architectures based on the HR-Net \citep{wang2020deep} for segmenting buildings of various shapes and sizes via aggregating the multi-scale features. GAMSNet \citep{lu2021gamsnet} used a multiscale residual framework based on U-Net \citep{ronneberger2015u} for road segmentation. Compared to GAMSNet, MAP-Net and MANet, where the authors followed the traditional methods for feature downsampling and self-attention-based multi-scale feature aggregation, our proposed architecture downsampled the feature maps using the DAMIP modules and aggregated the multi-scale features using the DAMSCA modules, developed on the prior generated probability maps of segmented regions of multiple-scales. Our method prevents heavy semantic information loss and improper attention to irrelevant regions of the feature maps compared to them.

\noindent \textbf{Multi-scale Supervision:} Multi-scale supervision or deep supervision \citep{lee2015deeply, li2018deep} increases the gradient flow by directly optimizing the hidden layers during backpropagation. It leads to faster model convergence as co-training the hidden layer companion objectives discover the complex internal latent space faster than optimizing the final layer. Multi-scale supervision has shown remarkable performance in image segmentation. Few notable pieces of research in the medical image segmentation are  Unet++ \citep{zhou2019unet++}, D3MS-Unet \citep{bose2022dense}. Multiscale aggregation FCN (MA-FCN) \citep{wei2019toward} had shown a decent performance on the WHU dataset by exploiting multiscale deep supervision along with polygon regularization. Even for segmentation of the building shadows over nearby areas, the authors in \citep{luo2020deeply} have developed a deeply supervised U-Net-based architecture by supervising the multi-scale probability maps generated by the decoder. Prior works only used supervision to increase the overall gradient flow. Moreover MAFCN \citep{wei2019toward} leverages deep supervision. Compared to MAFCN our proposed method is extensively supervised by multi-scale supervision and uses the generated segmentation maps of multiple resolutions to provide context-aware attention to the network.

\section{Methodology}

\subsection{Architecture of the proposed MSSDMPA-Net}


The network architecture of the proposed MSSDMPA-Net is illustrated in Fig \ref{fig2}. MSSDMPA-Net is a network consisting of \textbf{4 multiple} paths for encoding image features of various spatial resolutions. We denote $i$ as a single path where $i \epsilon \{1,2,3,4\}$. Each $i^{th}$ path consists of a dilated Single-Level Encoder, DAMIP, and DAMSCA module. Finally, the features of all the multiple paths are aggregated for input to the Decoder of MSSDAMPA-Net. We first highlight the four major components at any $i^{th}$ level given below, considering the image dimensions as channel $C=3$, height $H=512$, and width $W=512$, respectively, which are discussed in detail.
\begin{enumerate}
    \item Single level Encoder: Each block contains a feature encoder ($\mathcal{G}_i$) having repeated modules of dilated convolution and, finally, a DPMG $(\mathcal{F}_i)$ to extract semantic information at a specific scale. Overall, Four single-level encoder blocks are used in MSSDMPA-Net to extract the semantic features at various spatial resolutions. 
    \item DAMIP (${\mathcal{P}}_i$): It downsamples the spatial resolution of the input feature map without losing the semantic information by leveraging a novel index-pooling (${\mathcal{IP}}$) layer. Each of the three DAMIP modules in MSSDMPA-Net takes two inputs in parallel for attention-based learning, i.e., the input image feature map or the output feature of the previous DAMIP module and multi-path output probability map produced by the encoder. The next DAMIP module consumes the output feature map of the previous path DAMIP module.
    \item DAMSCA (${\mathcal{A}}_i$): Each of the four DAMSCA modules in MSSDMPA-Net performs novel attention-based learning on the Feature Encoder generated feature map using the DPMG Module generated output probability map of its corresponding level. Each DAMSCA module upsamples the input feature to a fixed higher spatial resolution, and finally, the outputs of all the DAMSCA modules are concatenated before processing by a decoder.
    \item Decoder (${\mathcal{D}}$): At the end, the decoder block is responsible for generating the output probability map.
\end{enumerate}

The input to the MSSDMPA-Net is an RGB image which is processed by the dilated convolutions of the Multi-Path Encoder to generate feature maps and probability maps of multiple-scales. These probability maps are used to downsample the spatial resolution of feature maps via the attention mechanism of the DAMIP module. Then the DAMSCA module upsamples the multi-scale features to a higher resolution via the module's novel attention mechanism using the multi-scale probabilty maps. Finally the high-level feature outputs of the DAMSCA modules are concatenated and passed through the Decoder Module to generate the ultimate segmentation map of the MSSDMPA-Net. The final decoder-produced segmentation map and four intermediate DPMG module-produced outputs are jointly optimized using multi-scale supervision.

\subsection{Details of the MSSDMPA-Net components}

\begin{figure*}[!htbp]
\centering
\includegraphics[width=\linewidth]{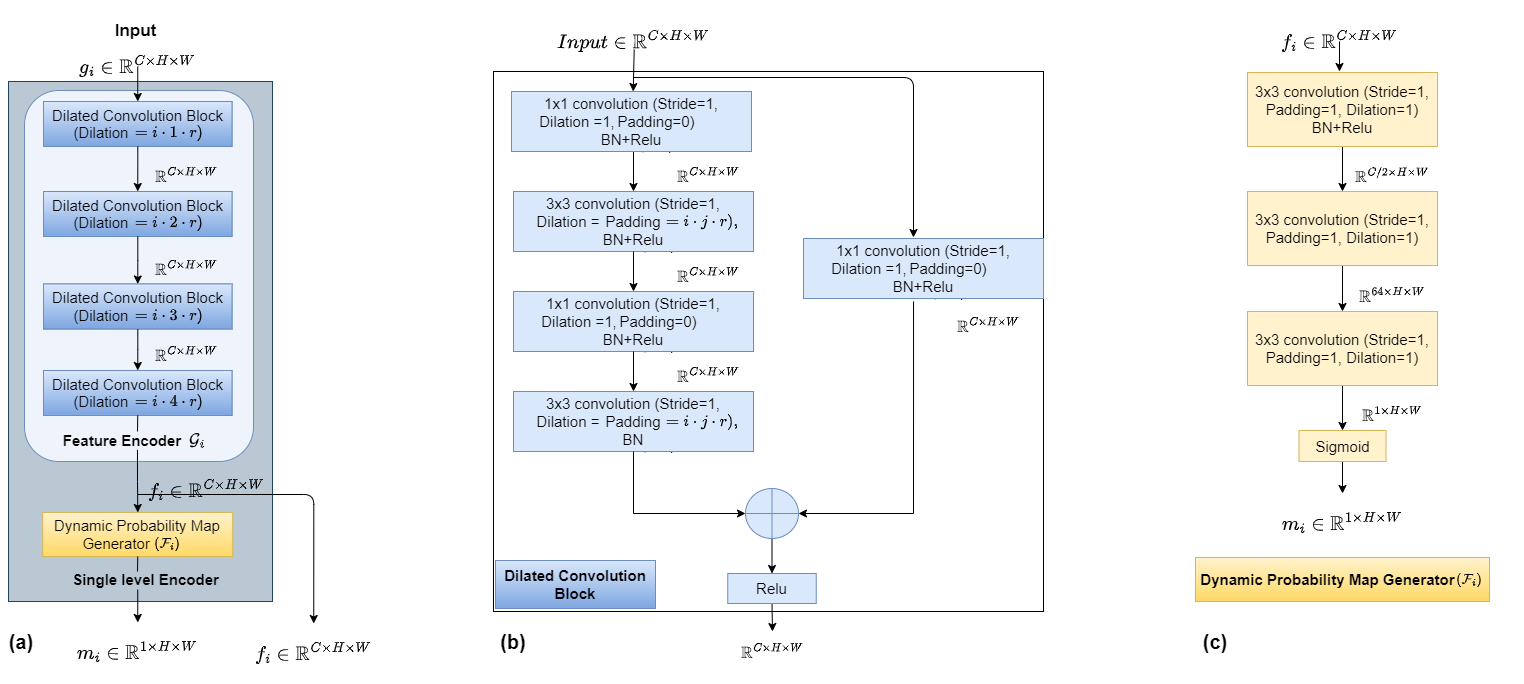}
\vspace*{-6mm}
\caption{The architecture of Single level Encoder and its constituents, namely, Dilated convolution block and DPMG ($\mathcal{F}_i$). Inside the single level encoder in (a), several dilated convolution blocks, shown in (b) are connected serially with an increasing dilation rate $i.j.r$ to construct the feature encoder, followed by a DPMG module in (c), where $j$ denotes the dilated convolution block number of the $i^{th}$ path with $r$ as constant. The DPMG block processes the encoded features from the feature encoder to generate the segmented probability map, which helps in attention-based learning.}
\label{fig3}
\end{figure*}

\subsubsection{\textbf{Single level Encoder}}
To capture the intrinsic semantic information for a larger receptive field without increasing computation cost, researchers have widely used dilated convolutions \citep{yu2015multi} in CNN. In contrast to regular convolution operation, dilated convolution also preserves the sharp geometric shapes and object contours in an image at various depths of a deep network \citep{zhou2018d, bose2022dense, yu2017dilated}. Mathematically, a dilated convolution kernel ${\theta_{k,d}}$ convolves an input feature map ${x_{in} \in \mathbb{R}^{C_{in} \times H_{in}\times W_{in}}}$ with a kernel size of $k$ and a dilation rate of ${d \in \mathbb{Z}^+}$. $k$ is effectively enlarged to ${k + (k-1)(d-1)}$ with a special case of $d=1$ in case of standard convolution.

\begin{equation}
    \label{eq:1}
     \hat f(x_{in})=\sum_{c=1}^{C_{in}} {\theta_{k,d} * x_{in}^c}
\end{equation}

where '$*$' denotes the convolution operation, and the dilated convolution on $x_{in}$ makes, ${\hat f:\mathbb{R}^{C_{in} \times H_{in}\times W_{in}}}$ ${\rightarrow \mathbb{R}^{C_{out} \times H_{out}\times W_{out}}}$ with $H_{in}, W_{in}, C_{in}$ are the input feature map height, width and channel dimensions, and $H_{out}, W_{out}, C_{out}$ are the dimensions for dilated convolved feature map.

Standard convolution operation with a constant dilation rate, $d=1$, fails to extract remotely-sensed tiny urban objects, especially the features of small buildings, and hazy road connections are completely lost in the smallest feature dimension of an encoder. Hence, we construct our Feature Encoder consisting of a series of Dilated Convolution Blocks Fig \ref{fig3}b having incremental dilation rates to preserve the sharp geometric characteristics of the building polygons and the road networks, which are extremely important for accurate segmentation. In designing each path of the multi-path encoder, responsible for processing feature maps of specific spatial resolution, we connect this feature encoder to a novel DPMG module for supervision. In the following sections, we will discuss the Feature Encoders consisting of the Dilated Convolution Blocks and the DPMG modules for each of the multiple paths.

\noindent \textbf{Feature Encoder:} Each dilated convolution block in the feature encoder in Fig. \ref{fig3}b uses an incremental dilation rate for the same path. Formally we define, the dilation factor of the $j^{th}$ dilated convolution block of the $i^{th}$ path in MSSDMPA-Net as $i.j.r$. We keep constant $r$ value, $r=1$ throughout the network and $i \epsilon\{1,\cdots, 4\}$ as we considered four paths in designing MSSDMPA-Net. We experimentally found that incrementing $j$ till $j=4$  gives the optimal solution. Further details on the maximum number of dilated convolution blocks in the feature encoder of a single path are available in the quantitative analysis section. Thus the dilation rates of the four consecutive dilated convolution blocks of the first path are $\{1,2,3,4\}$, then for the second path, the rates are $\{2,4,6,8\}$, the third path has the sequence of $\{3,6,9,12\}$, and finally, the dilation rates for the fourth path become $\{4,8,12,16\}$.  

Inside each dilated convolutional block in Fig. \ref{fig3}b, there exists two times repeated alternative of $1\times1$ and $3\times3$ convolution layers with dilation rate = $i.j.r$, padding = $i.j.r$, stride = $1$, Batch Normalization and Relu layer. Also, there is a residual connection \citep{he2016deep} using a $1\times1$ bottleneck convolution layer whose output is summed elementwise to the output of the final $3\times3$ convolution layer of the corresponding dilated convolutional block. The final summed output is forwarded through a non-linear Relu activation layer.


We define the feature encoder for the $i^{th}$ path of MSSDMPA-Net as $\mathcal{G}_i$ with the learnable parameters $W_{\mathcal{G}_i}$. Input to $\mathcal{G}_i$ is $g_i$ and the output is denoted as $f_i$. Thereby $f_i={\mathcal{G}_i}(g_i,W_{\mathcal{G}_i})$ for $i \epsilon\{1,\cdots,4\}$. The output feature dimensions of the four feature encoders operating at decrementing scales are $f_1 \epsilon \mathbb{R}^{C \times H/2 \times W/2}$, $f_2 \epsilon \mathbb{R}^{2C \times H/4 \times W/4}$, $f_3 \epsilon \mathbb{R}^{4C \times H/8 \times W/8}$, $f_4 \epsilon \mathbb{R}^{8C \times H/16 \times W/16}$. For example if $C=64, H=512, W=512$, then $f_i$ become, $f_1 \epsilon \mathbb{R}^{64 \times 256 \times 256}$, $f_2 \epsilon \mathbb{R}^{128 \times 128 \times 128}$, $f_3 \epsilon \mathbb{R}^{256 \times 64 \times 64}$, $f_4 \epsilon \mathbb{R}^{512 \times 32 \times 32}$.

\noindent \textbf{Dynamic Probability Map Generator (DPMG):}
The DPMG module $(\mathcal{F}_i)$ further processes the latent features ($f_i$) from the feature encoder to generate the segmented probability map ($m_i$), which helps in attention-based learning by the DAMIP module and the DAMSCA module. We define, $m_i=\mathcal{F}_i(f_i,W_{\mathcal{F}_i})$, where $W_{\mathcal{F}_i}$ is the learnable parameters of $\mathcal{F}_i$ for the $i^{th}$ path. Since, $f_i={\mathcal{G}_i}(g_i,W_{\mathcal{G}_i})$, thus, $m_i={\mathcal{F}_i}({\mathcal{G}_i}(g_i,W_{\mathcal{G}_i}),W_{\mathcal{F}_i})$. Using multi-scale supervision, DPMG-generated segmented map $m_i$ is directly optimized as companion objectives to the final decoder objective function to learn the intrinsic semantic structures of the complex urban area at an early stage. During training, $m_i$ is dynamically updated in each iteration; hence we define these maps as dynamic probability maps. As shown in Fig \ref{fig3}c, DPMG module consists of three consecutive units, comprising of a convolution layer with kernel size = $3 \times 3$, stride = 1, dilation rate = 1, padding = 1, Batch Normalization layer and a non-linear ReLU layer. At the end of the last convolution layer, a dense layer with a sigmoid activation function is used to squash the probability output in between [0, 1]. The dimensions of $m_i$ for four different paths in MSSDMPA-Net are,  $m_1 \epsilon \mathbb{R}^{1 \times 256 \times 256}$, $m_2 \epsilon \mathbb{R}^{1 \times 128 \times 128}$, $m_3 \epsilon \mathbb{R}^{1 \times 64 \times 64}$, $m_4 \epsilon \mathbb{R}^{1 \times 32 \times 32}$.

\subsubsection{\textbf{Dynamic Attention Map Guided Index Pooling (DAMIP)}}

\begin{figure*}
\centering
\includegraphics[width=\linewidth]{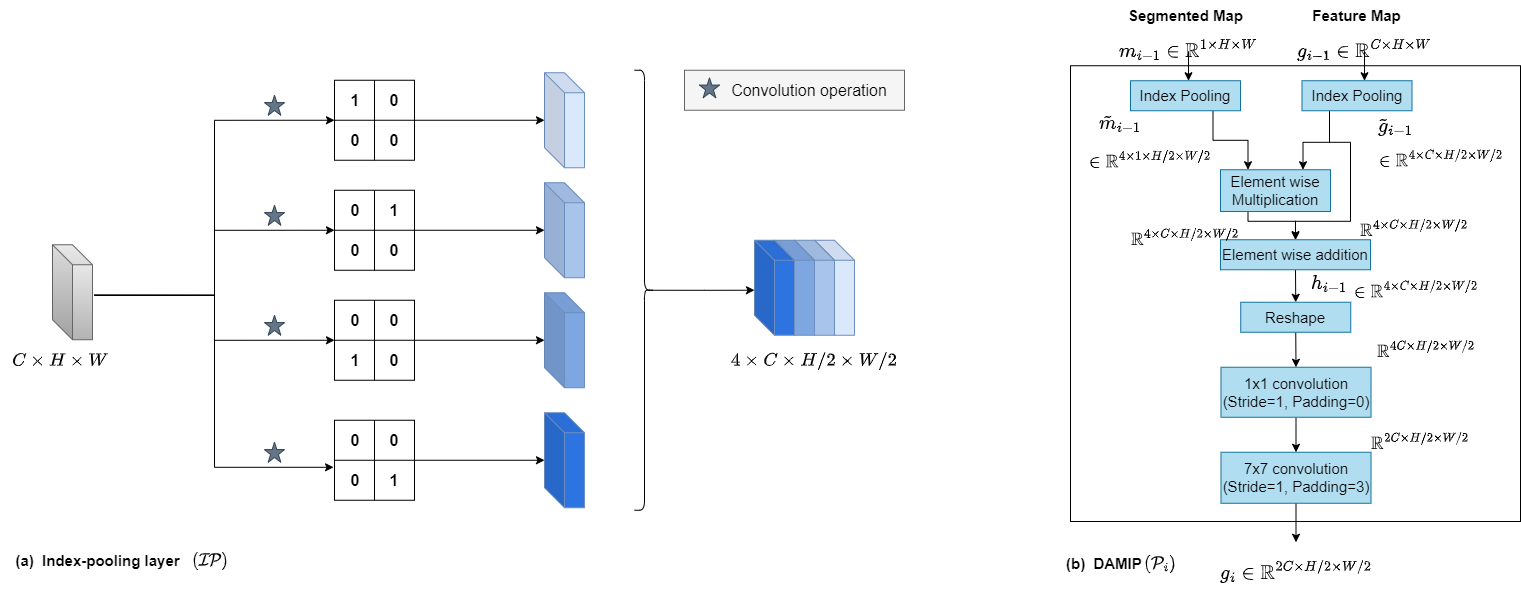}%
\vspace*{-7mm}
\caption{Illustration of the novel Index Pooling mechanism and its usage in DAMIP module $(\mathcal{P}_i)$ for the $i^{th}$ path of MSSDMPA-Net. (a) The Index pooling layer $(\mathcal{IP})$ downsamples the previous layer feature spatially by distributing its values across channel dimensions with binary kernels. Each kernel contains a single 'one' value at a unique position to mine the previous layer feature map value for a specific indexed location. For a $(2 \times 2)$ kernel with stride = 2, the Index Pooling layer reduces the spatial dimension by $(2,2)$ and increases the channel dimension by a magnitude of 4. Then, (b) using these spatially downsampled features, the DAMIP module performs attention-based learning to amplify salient information broadcasting.}
\label{fig4}
\end{figure*}

Using a novel index pooling mechanism, the DAMIP module $(\mathcal{P}_i)$ of the $i^{th}$ path in Fig. \ref{fig4}b first downsamples the DAMIP output $g_{i-1}$ and DPMG generated probability map $m_{i-1}$ from the previous path. It then amplifies the feature saliency with the downsampled map $m_{i-1}$. Finally, the amplified features are again downsampled in the channel dimensions using a bottleneck convolution layer followed by a $7 \times 7$ convolution layer to produce $g_i$. The advantage of using index pooling over traditional pooling mechanisms in CNN is that all the original values from the features and the probability map are distributed in their corresponding index positions while downsampling spatially. It helps in attention-based learning for downsampled encoded features with a lower memory footprint.

\noindent \textbf{Index Pooling:} The index pooling $(\mathcal{IP})$ layer with kernel size $k \times k$, downsamples a feature map into $k^2$ disjoint features, each containing the original feature map value for different index locations. To downsample a feature map using index pooling, first, we create $k^2$ kernels, each having a single `one' value at a unique index location from the set of $\{(0,0),\cdots, (k-1,k-1)\}$ positions and the rest are filled with `zero'. We convolve the feature map with this newly generated kernel set to obtain $k^2$ feature maps. Then, we concatenate these $k^2$ features and perform depthwise convolution to transfer the information along the channels. The relationship between the spatial dimensions of the input feature, $H_{in} \times W_{in}$ and output feature $H_{out} \times W_{out}$ is given in eq. \ref{pooling_eq}, where the convolution kernel size is $k \times k$, stride = $s$, dilation = $r$, padding = $p$. For zero padding ($p$ = 0), if we consider stride (s) = $k$, dilation (r) = 1, then the output feature dimension  $H_{out} \times W_{out}$ follows eq. \ref{pooling_eq1}. In the case of valid padding or $p=\frac{k\times ceil(\frac{H_{in}}{k})-H_{in}}{2}$, the dimensions follows according to eq.\ref{pooling_eq2}.

\begin{equation}
    \begin{array}{rcl}
H_{out} & = & floor(\frac{H_{in}+2 \times p-r \times (k-1)-1}{s}+1), \\
W_{out} & = & floor(\frac{W_{in}+2 \times p-r \times (k-1)-1}{s}+1) \\
\end{array}
\label{pooling_eq}
\end{equation}

\begin{equation}
H_{out} = floor(\frac{H_{in}}{k}), 
W_{out} = floor(\frac{W_{in}}{k}) \\
\label{pooling_eq1}
\end{equation}

\begin{equation}
H_{out} = ceil(\frac{H_{in}}{k}), 
W_{out} = ceil(\frac{W_{in}}{k}) \\
\label{pooling_eq2}
\end{equation}

The depthwise convolution is a convolutional operation where, a single convolutional filter is applied per each input channel. This is done by tuning the convolutional parameter groups as the number of input channels.

In MSSDMPA-Net, we downscale the input image with index pooling at a rate of two in each path, similar to the traditional CNN with $2\times 2$ pooling operation. We first generate four $2\times2$ kernels, where the first kernel has the value of `one' at the (0,0) index and the remaining filled with zero. Similarly, the second kernel has a value of `one' at index (0,1), the third kernel has a value of `one' at (1,0), and the fourth one has a value of `one' at index (1,1) with all the remaining indices in each kernel are filled with zero. As shown in Fig \ref{fig4}a, we convolve each of these kernels over the feature maps with a stride of 2. In doing so, we generate four feature maps with half the spatial resolution compared to the original feature maps. All the pixel information of the original feature maps is preserved in these four downsampled feature maps.        


The feature map $g_{i-1}$ is a 3-D tensor of shape $C\times H\times W$, whereas the probability map $m_{i-1}$ has a dimension of  $1\times H\times W$. The feature maps and the probability maps are passed through the index pooling layer to generate 4-D tensors of shape $4\times C\times \frac{H}{2}\times \frac{W}{2}$ for the index-pooled feature map  $(\tilde g_{i-1})$ and $4\times 1\times \frac{H}{2}\times \frac{W}{2}$ for the index-pooled probability map ($\tilde m_{i-1}$), respectively. Then we perform an element-wise multiplication between these two 4-D tensors $\tilde g_{i-1},\tilde m_{i-1}$ and add the dot product result with the input feature map $\tilde g_{i-1}$, to generate high-level informative features $h_{i-1}$.


\begin{equation}
    h_{i-1} = \mathcal{IP}({g_{i-1}}) \cdot \mathcal{IP}({m_{i-1}})+\mathcal{IP}({g_{i-1}})
\label{DAMIP_Eqn}
\end{equation}

Next, $h_{i-1} \epsilon \mathbb{R}^{4 \times C \times {H/2} \times {W/2}}$ is reshaped from a 4-D tensor to a 3-D one by making $h_{i-1} \epsilon \mathbb{R}^{4C \times {H/2} \times {W/2}}$. The channels are then downsampled from 4C to 2C by passing it through a $1\times1$ Convolution. Finally, these downsampled feature maps are processed by a $7\times7$ Convolution followed by BatchNorm and Relu activation. Thus, we produce attention-guided, highly informative downsampled feature maps $g_{i}$. According to Fig. \ref{fig2}, the first DAMIP module input $(g_1)$ is produced by passing the input image through a $7\times7$ Convolution layer with stride = 2 and padding = 3, followed by batch norm and Relu activation. Subsequent DAMIP module outputs from the remaining paths are denoted by $g_2,g_3,g_4$ which follows $g_{i} = {\mathcal{P}_i} (\{g_{i-1}, m_{i-1}\}, W_{\mathcal{P}_i})$, for $i > 1$ and $W_{\mathcal{P}_i}$ is the learnable parameters of $\mathcal{P}_i$ for the $i^{th}$ path. For example, if $C = 64$ in MSSDMPA-Net, then $g_1 \epsilon \mathbb{R}^{64 \times 256 \times 256}$, $g_2 \epsilon \mathbb{R}^{128 \times 128 \times 128}$, $g_3 \epsilon \mathbb{R}^{256 \times 64 \times 64}$, $g_4 \epsilon \mathbb{R}^{512 \times 32 \times 32}$.

\subsubsection{\textbf{Dynamic Attention Map Guided Spatial and Channel Attention (DAMSCA)}}

\begin{figure}
\centering
\includegraphics[height=3.25in,width=2.5in]{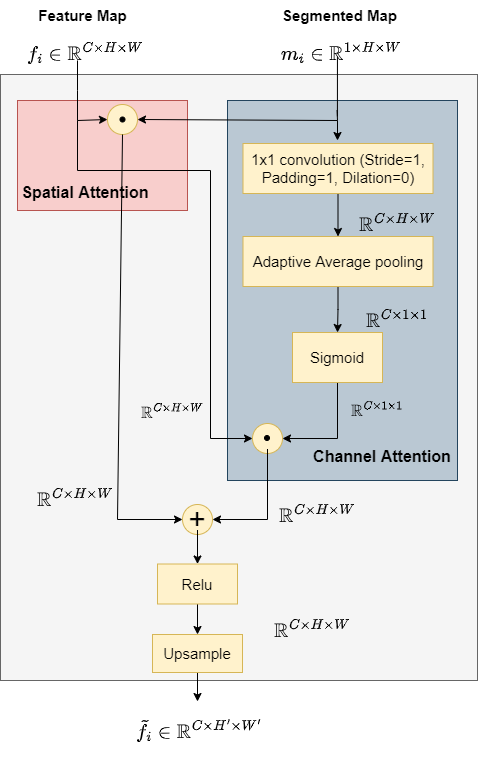}
\vspace*{-6mm}
\caption{Illustration of the novel spatio-spectral attention block, DAMSCA ($\mathcal{A}_i$). Using DPMG-generated segmented probability maps $(m_i)$, DAMSCA performs attention-learning in the spatial and channel dimensions. $(m_i)$ is directly applied to a multi-scale feature map $(f_i)$ for spatial attention. Whereas for channel attention, the probability map is convolved and average-pooled before performing channel-wise multiplication with $(f_i)$. Finally, spatial and channel attention outputs are added and upsampled to produce ($\tilde{f_i}$).}
\vspace*{-2mm}
\label{fig5}
\end{figure}

Existing attention mechanisms, CBAM \cite{woo2018cbam}, BAM \cite{park2018bam} and MTAN \cite{li2021multiattention} apply a processed version of the feature map as an attention mask to the same feature map itself. Without having the true guidance of the salient parts, the irrelevant parts of the feature map get amplified. In contrast, our DAMSCA module at the $i^{th}$ level applies a novel attention mechanism where the supervised probability maps $(m_i)$ from the $i^{th}$ level DPMG module act as the attention mask to the feature map output $(f_i)$ of the multi-path encoder's feature encoder $i^{th}$ level. It helps propagate correctly amplified feature details at each scale of the multi-path encoder.

As in Fig \ref{fig5}, $f_i$ and $m_i$ are the inputs to the $i^{th}$ path DAMSCA module and it produces the upsampled, correctly amplified salient feature $\tilde{f_i}$ as shown in Fig. \ref{fig5}. DAMSCA module has a spatial and channel attention part. For spatial attention, we perform a dot product between the input feature $f_i$ and the probability map $m_i$ to generate a spatially highlighted feature map, $f_i \cdot m_i$. Then for channel attention, we pass the probability map $m_i$ through a $1\times1$ convolution layer to increase the channel dimension from one to the number of input feature map channels $C$. We apply Global Average Pooling (GAP) over the output of the $1\times1$ convolution to reduce the spatial dimension from $H\times W$ to $1\times1$ before passing through a sigmoid activation function to scale the attention map in between [0, 1]. We multiply these channel-wise processed probability maps with the input feature maps to generate a channel highlighted feature map. Finally, the spatially and channel-wise highlighted feature maps are summed element-wise before passing through a Relu activation function. We upsample these high-level feature maps using a bilinear upsampler to a spatial dimension of $H'\times W'$.
We denote $\mathcal{A}_i$ as the DAMSCA module of $i^{th}$ path with learnable parameters $W_{\mathcal{A}_i}$. Then, we can write, $\tilde{f_i}={\mathcal{A}_i}(\{f_i,m_i\},W_{\mathcal{A}_i})$. Since all the $\tilde{f_i}$'s are upsampled to the same spatial dimension of $\frac{H}{2}\times\frac{W}{2}$ and we use $H=512, W =512$ in MSSDMPA-Net, hence the dimension of all $\tilde{f_i}$ becomes, $\tilde{f_i} \epsilon \mathbb{R}^{C_i' \times 256 \times 256}$, where $C_i'=2^{(i-1)}C$ is the number of channels in $\tilde{f_i}$. 

\subsubsection{\textbf{Decoder}}


The Decoder block ${\mathcal{D}}$, shown in \ref{fig2} starts with a transposed convolution layer of kernel size = $4\times4$ ,  stride = 2, padding = 1, and dilation rate = 1 to upsample the spatial resolution of input feature maps to $H\times W$ from $\frac{H}{2}\times\frac{W}{2}$ and to reduce the number of features in the channel dimension to $2C$ from $15C$. Next, three consecutive $3\times3$ convolution layers are used with stride = 1, padding = 1, and dilation rate = 1 to successively reduce the channel dimension from $2C$ to $C, \frac{C}{2}, 1$, respectively. A sigmoid activation to the end of the last convolution layer produces the segmented probability map, which is optimized along with DPMG-generated probability maps as part of the multi-scale supervision. The output feature maps $\tilde{f_1} \epsilon \mathbb{R}^{C \times H/2 \times W/2}$, $\tilde{f_2} \epsilon \mathbb{R}^{2C \times H/2 \times W/2}$, $\tilde{f_3} \epsilon \mathbb{R}^{4C \times H/2 \times W/2}$, $\tilde{f_4} \epsilon \mathbb{R}^{8C \times H/2 \times W/2}$ from the four DAMSCA modules corresponding to each path of the MSSDMPA-Net are concatenated in channel dimension, $f_{dec} = (\tilde{f_1}||\tilde{f_2}||\tilde{f_3}||\tilde{f_4})$, producing $f_{dec} \epsilon \mathbb{R}^{15C \times H/2 \times W/2}$, where $||$ denotes the concatenation operation. The Decoder then generates the ultimate segmentation map of MSSDMPA-Net, $m_{out}$ from the $f_{dec}$ features, where $m_{out} \epsilon \mathbb{R}^{1\times H\times W}$.  For example if $C=64$ and $H=512, W=512$, then $f_{dec} \epsilon \mathbb{R}^{960 \times 256 \times 256}$. Decoder block processes $f_{dec}$ to generate the final segmentation map $m_{out} \epsilon \mathbb{R}^{1 \times 512 \times 512}$.

\subsection{Loss Function}
Noise-robust dice loss \citep{wang2020noise} alleviates the class imbalance problem between the foreground and background classes and is more tolerant against noisy labels than simple dice loss. In order to make MSSDMPA-Net robust towards noisy labels, we use $\mathcal{L}$, a combination of noise-robust dice loss $\mathcal{L}_{NR-Dice}$ along with binary cross-entropy loss $\mathcal{L}_{BCE}$ to optimize each of the multi-scale generated probability maps,

\begin{align}
\mathcal{L}_{BCE}=-\sum_{p,q}[s_{p,q}\log(t_{p,q})+(1-s_{p,q})\log(1-t_{p,q})]    \label{bce}
\end{align}

\begin{equation}
\mathcal{L}_{NR-Dice} = \frac{\sum_{p,q}|t_{p,q}-s_{p,q}|^\gamma}{\sum_{p,q}t_{p,q}^2 + \sum_{p,q}s_{p,q}^2+\epsilon}      \label{nrdice}
\end{equation}

\begin{equation}
\mathcal{L}= \mathcal{L}_{BCE}+\mathcal{L}_{NR-Dice}   \label{totloss}
\end{equation}

where, $t_{p,q}$ indicates the pixel values of the predicted image, and $s_{p,q}$ indicates the corresponding pixel values of the ground truth for $(p,q)$ pixel index. $\gamma \epsilon [1.0, 2.0]$ is a hyperparameter for generalization, and $\epsilon$, is a small number used to avoid ambiguity for all negative cases. We have used $\gamma=1.5$ and $\epsilon=10^{-5}$ similar to \citep{wang2020noise}. Using multi-scale supervision, we optimize four DPMG-generated probability maps, $m_1 \epsilon \mathbb{R}^{1 \times 256 \times 256}$, $m_2 \epsilon \mathbb{R}^{1 \times 128 \times 128}$, $m_3 \epsilon \mathbb{R}^{1 \times 64 \times 64}$, $m_4 \epsilon \mathbb{R}^{1 \times 32 \times 32}$ and the final decoder predicted probability map $m_{out} \epsilon \mathbb{R}^{1 \times 512 \times 512}$, which further help to generate refined attention maps for the DAMIP and DAMSCA modules at each step of the iteration. The ground truth map $\hat y \epsilon \mathbb{R}^{1 \times 512 \times 512}$ has the same spatial dimension as the input image in MSSDMPA-Net. Thus to supervise the DPMG-generated probability maps, we downsample the ground truth map to the same sizes of $m_i$, reducing the ground truth dimension as $\hat y_1 \epsilon \mathbb{R}^{1 \times 256 \times 256}$, $\hat y_2 \epsilon \mathbb{R}^{1 \times 128 \times 128}$, $\hat y_3 \epsilon \mathbb{R}^{1 \times 64 \times 64}$, $\hat y_4 \epsilon \mathbb{R}^{1 \times 32 \times 32}$. Finally, the total loss function becomes:

\begin{equation}
    \mathcal{L}_{total} = \sum_{l=1}^{L}{\mathcal{L}(m_i, \hat y_i)}+\mathcal{L}(m_{out},\hat y)
    \label{total_loss}
\end{equation}

Where, $L$ is the number of paths in the multi-path encoder for multi-scale supervision. In this model, $L=4$.

\section{Implementation details}

\subsection{Dataset description}

To evaluate the proposed model, we have extensively investigated the performance of the proposed model on seven publicly available datasets. Among the seven datasets, three focus on road segmentation while the other three deal with building segmentation from satellite imagery.

\subsubsection{Porto dataset \cite{wu2020deepdualmapper}}

This is a single aerial image dataset with a width of 15.447 km and 13.538 km taken from satellites 
with a resolution of 0.90 m/pixel. The size of the whole image is $14388\times16747\times3$. We randomly cropped 80\% of the image region as the training region and the remaining 20\% region as the testing region and performed a 5-fold cross-validation.

\subsubsection{Shanghai dataset \cite{wu2020deepdualmapper}}

The original image resolution of this dataset is 1.02 m/pixel. The size of the whole image is $12500\times20000\times3$. We processed this dataset similar to the Porto dataset for training and testing purposes. 

\subsubsection{Massachusetts road segmentation dataset \citep{mnih2013machine}}

This aerial dataset consists of 1171 images of Massachusetts state. The size of each image is $1500 \times 1500$ pixels with a resolution 1m/pixel, covering an area of 2.25 square kilometers. 
The dataset is divided into 1108 training images, 14 validation images, and 49 testing images. 

\subsubsection{Massachusetts building segmentation dataset \cite{mnih2013machine}}

This dataset consists of 151 aerial images, each having a size of $1500\times1500$ with a resolution of 1m/pixel. The training, validation, and testing sets have 137, 4, and 10 images, respectively.

\subsubsection{Synthinel dataset \cite{kong2020synthinel}}

This is synthetic imagery dataset constructed using CityEngine software. This virtual imagery dataset contains nine different city styles (a, b, c, d, e, f, g, h, i), with each image having a size of $572\times572$ pixels with resolution of 0.3m/pixel. We have adopted eight city styles for our experiments: a, b, c, d, e, g, h, i. For each city style, we have randomly chosen 80\% of images for training and the remaining 20\% for testing. Each image has been resized into an image of shape $512 \times 512$. We performed 5-fold cross-validation experiments.



\subsubsection{WHU Satellite I dataset \cite{ji2018fully}}

This dataset is collected from 51 cities using various remote sensing resources. The number of images in the dataset is 204, with the size and resolution of each image being $512\times512$ and 0.3-2.5 m, respectively. Each unique-styled city has four images. Thus for performing the experiments, we have considered three images as training and the remaining image as testing for each city style. We then applied 4-fold cross-validation strategy.

\subsubsection{WHU Ariel Imagery dataset \cite{ji2018fully}}

This dataset consists of more than 187000 buildings, covering an area of 450 $km^2$ and having a ground resolution of 30 cm. The RGB images have a size of $512\times512$ pixels.
It consists of 8188 tiles of images which are divided into 4736, 1036, and 2416 tiles for training, validation, and testing, respectively.

\begin{figure*}[htbp]
\centering
\includegraphics[width=5in]{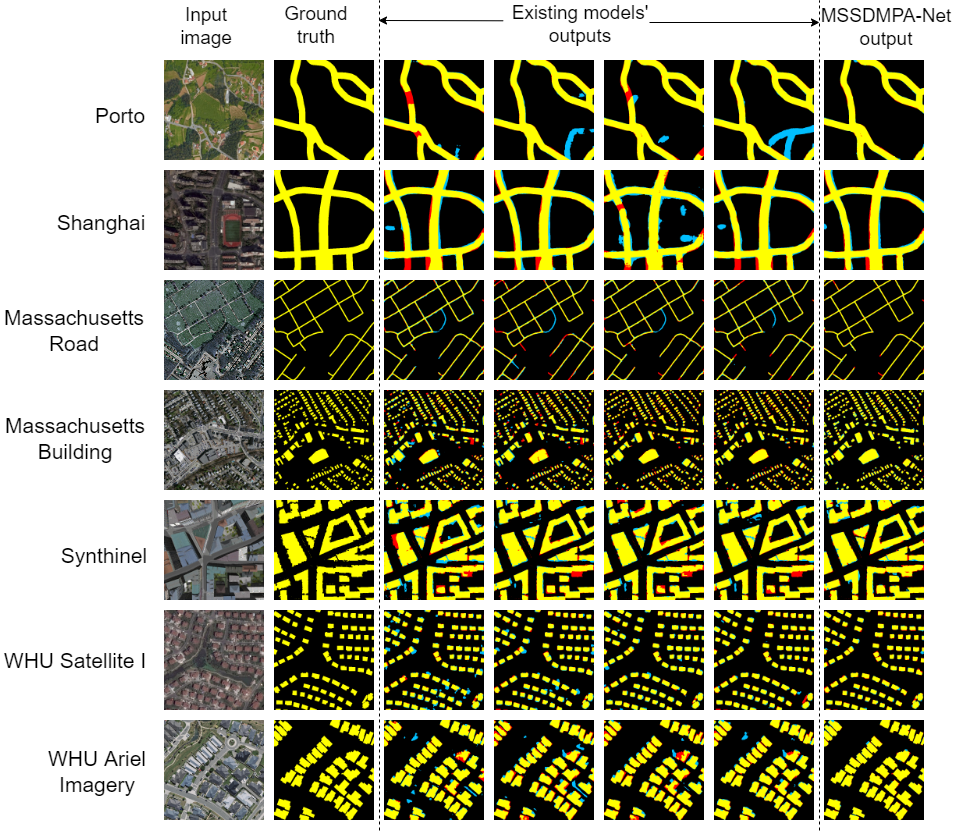}%
\hfil
\vspace*{-5mm}
\caption{The figure shows the output images of the existing models and MSSDMPA-Net over Porto, Shanghai, Massachusetts Road, Massachusetts Building, Synthinel, WHU Satellite I and WHU Ariel Imagery datasets. The first and second columns show the input images and ground truth images respectively. From the third column to the sixth column, the outputs of the existing models have been shown. The seventh column presents the output of MSSDMPA-Net. The existing models are: For Porto dataset: 1. Deep dual mapper, 2. Linknet, 3. Unet++ (DS), 4. Dlinknet; Shanghai dataset: 1. Deep dual mapper, 2. Linknet, 3. Unet++ (DS), 4. 1-D Decoder; Massachusetts Road: 1. ASPN Net, 2. HRNet, 3. Unet++ (DS), 4. Dlinknet; Massachusetts Building: 1. BRRNet, 2. Self cascaded CNN Resnet, 3. EU-Net, 4. ENRU-Net; Synthinel: 1. DeeplabV3+, 2. PSP-Net, 3. Dlinknet, 4. Linknet; WHU Satellite I: 1. Linknet, 2. Dlinknet, 3. DeeplabV3+, 4. PSP-Net and WHU Ariel Imagery: 1. Res2-Unet, 2. MSCRF, 3. MAPNet, 4. MAFCN respectively.}
\vspace*{-2mm}
\label{fig6}
\end{figure*}

\begin{figure*}[h]
\centering
\includegraphics[width=5in]{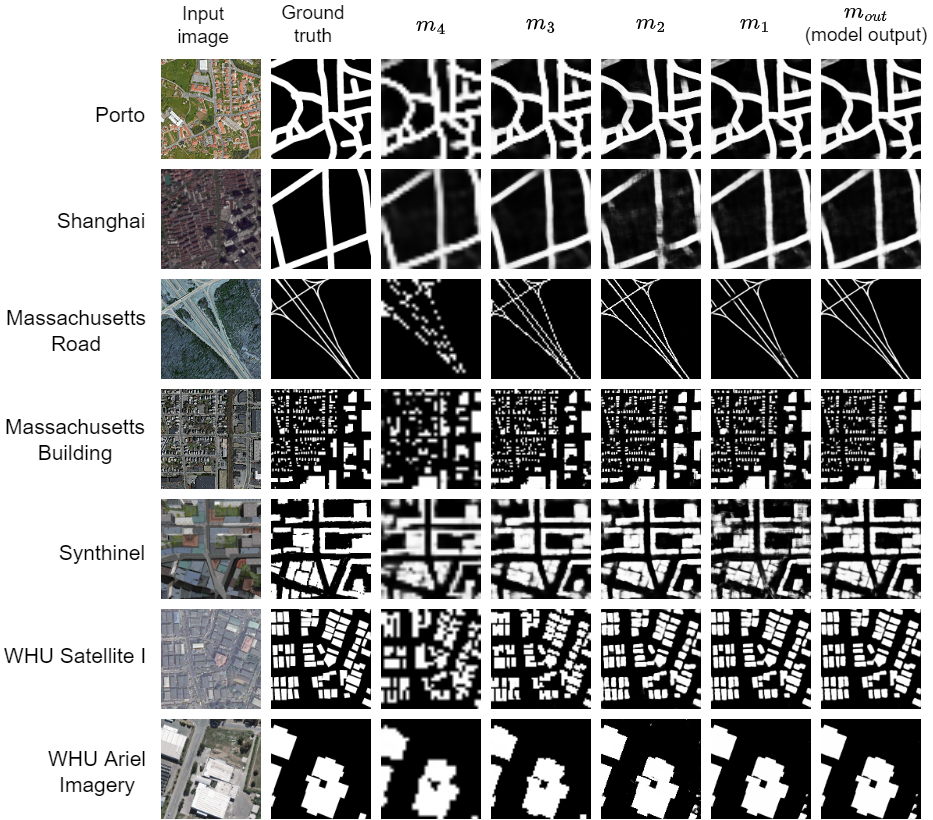}%
\hfil
\vspace*{-3mm}
\caption{The figure shows the segmentation probability maps (grayscale images) of the MSSDMPA-Net of Porto, Shanghai, Massachusetts Road, Massachusetts Building, Synthinel, WHU Satellite I and WHU Ariel Imagery datasets respectively. The first and second columns show the input images and ground truth images respectively. From the third column to the seventh column, the segmentation probability maps are presented from lowest resolution to highest resolutions ($32 \times 32, 64 \times 64, 128 \times 128, 256 \times 256$ and $512 \times 512$). The seventh column presents the model output.}
\label{fig7}
\end{figure*}

\subsection{Evaluation metrics}








The segmentation performance of our proposed method is evaluated using the Sorensen Dice Coefficient or F1, Intersection over Union (IoU), Precision, and Recall as the primary metrics for evaluation.

The IoU is expressed as the ratio between the overlapped region and the entire region covered by the ground truth and the predicted output. The F1 Score is used to measure the similarity between the ground truth and the predicted output extensively used in segmentation tasks.

\begin{equation}
\label{eq:1}
IoU = \frac{TP}{TP+FP+FN}
\end{equation}

\begin{equation}
\label{eq:2}
F1-Score = \frac{2\times TP}{2\times TP+FP+FN}
\end{equation}

$TP$, $TN$, $FP$ and $FN$ are True Positive, True Negative, False Positive, and False Negatives respectively.

The ratio between the true positive and the total positively predicted output is Precision. The ratio between the true positive and the total positively labeled ground truth is Recall.

\subsection{Experimental setup}

For training on these datasets, augmentations applied are random rotation by 90 degrees, random hue-saturation value shifting, random shifting with rotation from -90 degrees to +90 degrees, and random vertical and horizontal flips with a probability of 0.5. In addition, patches of size 512 × 512 are generated via a sliding window mechanism with no overlap between consecutive patches over the training, validation, and test sets for Porto, Shanghai, Massachusetts road segmentation, and Massachusetts building segmentation dataset.

We initially trained our proposed model for 60 epochs at a learning rate of 0.001 and then again trained our model for another 60 epochs at a learning rate of 0.0001 and finally trained our model for another 20 epochs at a learning rate of 0.00001 to complete the training procedure. We used the Adam\citep{kingma2014adam} optimizer to optimize our proposed model. We kept both the training and testing batch size fixed at 64. We train our model using Pytorch deep learning software with two units of 16 GB NVIDIA 2080 Ti GPU and 64 GB RAM.    

\section{Results and comparative analysis}

\begin{figure}
\centering
\includegraphics[width=\linewidth]{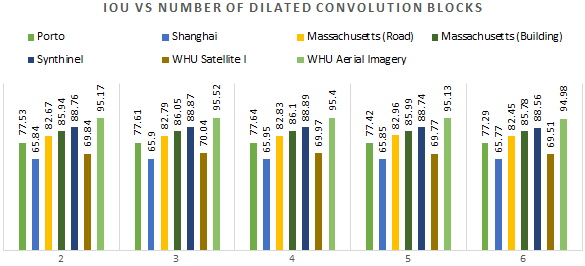}
\hfil
\caption{Performance of MSSDMAP-Net in terms of IoU with variation of number of Dilated Convolution blocks on different road and building segmentation datasets. }
\label{fig8}
\end{figure}

\subsection{Qualitative analysis}
Figure \ref{fig6} shows the input image, ground truth, and output maps of each dataset's proposed and existing state-of-the-art models. The yellow region represents the true positive (TP), the blue region represents the false positive (FP), the red region represents the false negative (FN), and the black region represents the true negative (TN). The output images of the proposed model, the Porto, Shanghai Massachusetts road, Synthinel, and WHU Ariel Imagery, have huge visual improvement compared to the existing models. In other datasets, Massachusetts building and WHU satellite I datasets, the proposed model outperforms all other existing state-of-the-art models. In Porto and Massachusetts road datasets, the existing models wrongly classify additional road, and in Synthinel, Satellite I, and WHU Ariel Imagery datasets, the existing models fail to segregate the buildings, especially the boundary regions, properly. The proposed model shows improved results in both road and building datasets. 

Figure \ref{fig7} shows the segmentation probability maps of the proposed model on the seven datasets. The segmentation probability maps are shown in order from lowest resolution to highest resolution. $m_4$, $m_3$, $m_2$ and $m_1$ have resolutions of $32 \times 32$, $64 \times 64$, $128 \times 128$ and $256 \times 256$ respectively. $m_4$, $m_3$, $m_2$ and $m_1$ are the output of the Multi-Path Encoder module. These images are directly used in the DAMIP and DAMSCA modules to produce the final output. $m_{out}$ is the model output with a resolution of $512 \times 512$. As deep supervision is used using these segmentation probability maps, thus these images are an approximation of the ground truth. From the images, we can see that $m_4$ is the worst approximation due to the lowest resolution. $m_{out}$ or the model output is overall the best approximation due to the highest resolution.

\subsection{Quantitative analysis}
We have compared the IoU segmentation score on different datasets to obtain the optimal number of Dilated Convolution blocks in the encoder. Fig \ref{fig8} depicts the variation of IoU with the number of Dilated Convolution blocks, where it is evident that 4 Dilated Convolution blocks will yield the best result. Hence we applied four blocks in our model.

\subsubsection{Results on road segmentation datasets}


\begin{table*}[htbp]
  \centering
  \caption{Performances of the proposed and existing models on road segmentation datasets}
    \vspace*{-3mm}
   \scalebox{0.7}{
    \begin{tabular}{|c|cccc|cccc|cccc|}
    \hline
    \textbf{Dataset} & \multicolumn{4}{c|}{\textbf{Porto}}            & \multicolumn{4}{c|}{\textbf{Shanghai}}   & \multicolumn{4}{c|}{\textbf{Massachusetts Road}}\\
    \hline
    \textbf{Methods} & {\textbf{Recall}} & {\textbf{Precision}} & {\textbf{IoU}} & {\textbf{F1 Score}} & {\textbf{Recall}} & {\textbf{Precision}} & {\textbf{IoU}} & {\textbf{F1 Score}} & {\textbf{Recall}} & {\textbf{Precision}} & {\textbf{IoU}} & {\textbf{F1 Score}}\\
    \hline
    ASPN Net \citep{shamsolmoali2020road} & 82.45 & 84.14 & 71.48 & 83.29 & 70.22 & 81.65 & 60.64 & 75.50 & 88.14 & 87.29 & 78.86 & 88.18  \\
    \hline
    Unet \cite{ronneberger2015u}   &   79.12    &  81.29    & 63.57  & 77.98   &    67.35   &    75.83   & 50.28  & 64.61 & 83.72 & 81.33 & 70.21 & 82.50\\
    \hline
    U-Net++ \cite{zhou2019unet++} &    83.47     &   83.67    & 70.34 &   83.31   &     71.27  &   79.17    & 58.11 & 73.32 & 86.95 & 85.52 & 75.77 & 86.22\\
    \hline
    U-Net++ (DS) \cite{zhou2019unet++} & 83.71 & 83.3  & 71.37 & 83.40 & 71.45 & 80.22 & 59.91 & 75.76 & 87.21 & 85.93 & 76.31 & 86.56 \\
    \hline
    SegNet \cite{badrinarayanan2017segnet}   &    79.10   &   80.89    &   61.14    &    76.34   &   66.18    &    73.49   &    48.19   & 62.87 & 82.01 & 81.29 & 68.98 & 81.64\\
    \hline
    Hrnet \cite{wang2020deep} &  80.32 & 83.24 & 69.49 & 81.52 & 68.90  & 81.61 & 59.03 & 74.51 & 87.09 & 86.17 & 76.39 & 86.62 \\
    \hline
    Linknet \cite{chaurasia2017linknet} & 92.10  & 83.18 & 83.75 & 71.48 &  69.70  & 83.50  & 60.81 & 75.86 & 86.59 & 85.1 & 75.18  & 85.83 \\
    \hline
    Dlinknet \cite{zhou2018d} & 92.16 & 83.33 & 82.99 & 71.28 &  70.70  & 81.11 & 60.04 & 75.45 & 86.63 & 85.37 & 75.42 & 85.99  \\
    \hline
    MAFCN \cite{wei2019toward} &      81.87   &   83.18    & 70.05 &   82.31 & 69.21 & 70.39 & 52.9  & 69.96 & 85.22 & 86.01 & 74.84 & 85.61 \\
    \hline
    1D Decoder \cite{sun2019leveraging} &  79.27   &    81.89   & 67.15  & 80.10    &   70.14   &   82.16    & 60.59    & 75.46 & 84.05 & 81.87 & 70.85 & 82.94\\
    \hline
    DeepDual Mapper \cite{wu2020deepdualmapper} &     83.87  &  83.95  & 71.7 & 83.5   &    71.63  &  83.82  & 63.4  & 77.6 & 84.23 & 83.31 & 72.07 & 83.77 \\
    \hline
    \textbf{MSSDMPA-Net}   & \textbf{86.75} & \textbf{88.23} & \textbf{77.64} & \textbf{87.48}      &   \textbf{73.38}    &    \textbf{85.48}   &   \textbf{65.95}    & \textbf{79.50}  & \textbf{91.11} & \textbf{90.11} & \textbf{82.83} & \textbf{90.61}  \\
    \hline
    \end{tabular}%
    }
  \label{tab1}%
\end{table*}%

\begin{table*}[h]
  \centering
  \caption{Performances of the proposed and existing models on building segmentation datasets}
  \vspace*{-3mm}
   \scalebox{0.7}{
    \begin{tabular}{|c|cccc|cccc|cccc|cccc|}
    \hline
    \textbf{Dataset} & \multicolumn{4}{c|}{\textbf{Massachusetts Building}} &
    \multicolumn{4}{c|}{\textbf{Synthinel}}        & \multicolumn{4}{c|}{\textbf{WHU Satellite I}}   & \multicolumn{4}{c|}{\textbf{WHU Ariel Imagery}}\\
    \hline
    \textbf{Methods} & \textbf{Recall}  & \textbf{Precision} & \textbf{IoU}   & \textbf{F1 Score} & \textbf{Recall}  & \textbf{Precision} & \textbf{IoU}   & \textbf{F1 Score} & \textbf{Recall}  & \textbf{Precision} & \textbf{IoU}   & \textbf{F1 Score} & \textbf{Recall}  & \textbf{Precision} & \textbf{IoU}   & \textbf{F1 Score}\\
    \hline
    Unet \cite{ronneberger2015u} & 80.23 & 81.48 & 67.85 & 80.85 & 87.17 & 81.11 & 73.41 &   85.15 & 72.42    &   69.48    &  56.82     & 71.13  & 91.4  & 94.5 & 86.80  & 92.92 \\
    \hline
    U-Net++ \cite{zhou2019unet++} & 80.15    &   83.67    &     69.34  & 81.56 & 85.22 & 71.19 & 62.27 & 77.46  & 79.64 & 70.11 & 59.07 & 74.46  & 92.34 & 95.69 & 88.64 & 93.98 \\
    \hline
    U-Net++ (DS) \cite{zhou2019unet++}  &    79.87   &  84.10     &   69.79    & 81.79 & 87.12 & 77.59 & 68.98 & 82.01  & 79.09 & 70.06 & 58.49 & 74.41 & 93.89 & 94.56 & 89.08 & 94.22 \\
    \hline
    Hrnet \cite{wang2020deep} &    80.26   &  83.67     &    68.12   & 81.31 & 86.32 & 78.51 & 69.60  & 82.35 & 74.22 & 72.00  & 57.24 & 73.23  & 88.11 & 92.24 & 82.02 & 90.12\\
    \hline
    Linknet \cite{chaurasia2017linknet} &  81.47    &   84.53    &   71.59    &  82.67 & 90.28 & 82.30  & 75.68 & 85.98 & 80.08 & 73.82 & 62.24 & 76.98 & 87.40 & 86.75 & 77.16 & 87.07 \\
    \hline
    Dlinknet \cite{zhou2018d}  &   82.27    &  84.79    &  72.26  &  83.73 & 89.88 & 82.94 & 75.85 & 86.39 & 76.88 & 76.24 & 61.87 & 76.46 & 88.35 & 87.10 & 78.12 & 87.72\\
    \hline
    MAFCN \cite{wei2019toward}   &     78.24    &  81.29  & 65.44 & 79.47  &   89.54 &   87.63   &  79.49  &   88.57  & 81.46 & 69.55 & 59.82 & 75.13 & 95.20  & 95.10 & 90.74 & 95.15\\
    \hline
    FCN8s \citep{long2015fully}  & 69.89 & 72.58 & 56.64 & 71.58 & 87.28 & 83.92 & 74.93 &   85.61 &  73.43   &   70.21    &   57.18    & 72.39 & 86.76 & 85.11 & 75.33 & 85.92 \\
    \hline
    Segnet \cite{badrinarayanan2017segnet} & 80.27  & 82.19  & 67.42 & 81.34 & 89.39 & 82.65 & 75.36 &  85.82      &   71.55    &    68.98   &    55.79   & 70.57 & 90.75 & 94.01 & 85.78 & 92.35\\
    \hline
    PSP-Net \citep{zhao2017pyramid} &  71.68 & 75.24 & 59.34 & 73.14 & 90.51  & 86.80  & 79.12 &    88.36      &  74.53     &    71.29   &    60.26   & 73.72 & 92.31 & 95.13 & 88.13 & 93.69\\
    \hline
    DeeplabV3+ \citep{chen2017rethinking}  & 72.90 & 76.86 & 60.85 & 75.32 & 90.49 & 86.54 & 79.47 &   88.43      &   78.34    &    74.23   &   61.37    & 75.91 & 87.88 & 88.38 & 78.77 & 88.13 \\
    \hline
    
    EU-Net \cite{kang2019eu} & 83.40  & 86.70  & 73.93 & 85.01 & 85.33 & 86.16 & 75.04 & 85.74 & 72.40 & 71.99 & 56.48 & 72.19 & 87.60 & 87.33 & 77.72 & 87.46 \\
    \hline

    BRRNet \cite{shao2020brrnet} &       84.44    &   86.45    & 74.46 & 85.36 & 87.70 & 86.49 & 77.13 & 87.09 & 74.21 & 77.19 & 60.86 & 75.67 & 88.12 & 87.99 & 78.66 & 88.05 \\
    \hline
    \begin{tabular}[c]{@{}l@{}} SC-CNN \cite{liu2018semantic}\end{tabular} &     84.32   & 86.13    & 74.34 & 85.58 & 88.68 & 88.20 & 79.27 & 88.44 & 75.42 & 75.99 & 60.91 & 75.70 & 88.39 & 89.22 & 79.86 & 88.80 \\
    \hline
    ENRU-Net \cite{wang2020automatic} &    84.26   &    85.06  & 73.02 & 84.41 & 87.52 & 87.45 & 77.75 & 87.48 & 74.71 & 74.30 & 59.37 & 74.50 & 89.24 & 89.69 & 80.94 & 89.46\\
    \hline
    MAPNet \cite{zhu2020map} & 85.30 & 83.63 & 73.09 & 84.46 & 89.71 & 90.55 & 82.03 & 90.13 & 77.49 & 77.19 & 63.05 & 77.34 & 95.62 & 94.81 & 90.86 & 95.21 \\
    \hline
    MSCRF \cite{zhu2020building} & 84.82 & 83.17 & 72.39 & 83.98 & 88.41 & 88.33 & 79.16 & 88.36 & 75.95 & 76.80 & 61.77 & 76.37 & 96.47 & 95.07 & 91.99 & 95.76 \\
    \hline
    Res2-Unet \cite{chen2022res2} & 86.20 & 85.55 & 75.24 & 85.87 & 87.56 & 88.11 & 78.30 & 87.83 & 74.99 & 76.24 & 60.78 & 75.61 & 96.57 & 95.99 & 92.83 & 96.28 \\
    \hline
    \textbf{MSSDMPA-Net}   & \textbf{92.72} & \textbf{92.34} & \textbf{86.10} & \textbf{92.53}   & \textbf{94.16} & \textbf{94.10} & \textbf{88.89} & \textbf{94.11} & \textbf{84.60} & \textbf{80.20} & \textbf{69.97} & \textbf{82.33}  & \textbf{97.50} & \textbf{97.82} & \textbf{95.40} & \textbf{97.65}\\
    \hline
    \end{tabular}%
    }
  \label{tab2}%
\end{table*}%

\begin{table*}[htbp]
  \centering
  \caption{Performances of the Ablation models on road segmentation datasets}
    \vspace*{-3mm}
  \scalebox{0.7}{
    \begin{tabular}{|c|c|cccc|cccc|cccc|}
    \hline
    \multicolumn{2}{|c|}{\textbf{Dataset}} & \multicolumn{4}{c|}{\textbf{Porto}}            & \multicolumn{4}{c|}{\textbf{Shanghai}} & \multicolumn{4}{c|}{\textbf{Massachusetts Road}} \\
    \hline
    \multicolumn{2}{|c|}{\textbf{Methods}} & {\textbf{Recall}} & {\textbf{Precision}} & {\textbf{IoU}} & {\textbf{F1 Score}} & {\textbf{Recall}} & {\textbf{Precision}} & {\textbf{IoU}} & {\textbf{F1 Score}} & {\textbf{Recall}} & {\textbf{Precision}} & {\textbf{IoU}} & {\textbf{F1 Score}} \\
    \hline
    Without
    & LIP \cite{gao2019lip} & 84.11 & 86.69 & 74.48 & 85.38 & 71.27 & 82.49 & 61.90 & 76.47 & 88.69 & 88.33 & 79.38 & 88.51\\
    \cline{2-14} DAMIP & Maxpool    &  83.82     &   86.36    & 74.25 &  85.15 & 70.97 & 82.92 & 61.51 & 76.25 &  89.43 & 88.16 & 79.84 & 88.79 \\
    \cline{2-14} & Avgpool & 83.53 & 85.76 & 73.27 & 84.69 & 70.37 & 82.01 & 60.50 & 75.59 & 89.08 & 88.03 & 79.45 & 88.55 \\
    \cline{2-14} & Stochastic pool \cite{zeiler2013stochastic} & 83.37 & 85.43 & 72.96 & 84.46 &  70.24 & 81.83 & 60.39 & 75.45  & 88.95 & 87.87 & 79.22 & 88.41   \\
    \hline
    \multicolumn{2}{|c|}{Without Deep Supervision} & 81.76 & 83.90  & 72.37 & 83.08 & 68.55 & 80.13 & 58.62 & 73.93 & 86.76 & 85.78 & 75.84 & 86.26\\
    \hline
    \multicolumn{2}{|c|}{Without Dilation}  &    84.19   &   87.60    &   75.16    &   85.92  & 71.53 & 82.40 & 62.29 & 76.72   & 89.12 & 88.54 & 79.89 & 88.82\\
    \hline
    \multicolumn{2}{|c|}{Without DAMSCABlock}  & 83.81 & 85.46 & 73.62 & 84.73 & 71.15 & 82.16 & 61.31 & 76.13  & 88.96 & 88.23 & 78.71 & 88.09\\
    \hline
    Different self  & SAGAN \cite{zhang2019self} & 84.76 & 88.56 & 76.67 & 86.76 & 72.23 & 83.42 & 63.05  & 77.33 & 90.24 & 88.78 & 80.57 & 89.24\\
    \cline{2-14} attention pluggins & CBAM \cite{woo2018cbam}  & 84.59 & 88.30 & 76.52 & 86.67 & 71.91 & 83.01 & 62.49 & 77.00 & 89.12 & 88.49 & 79.86 & 88.80\\
    \cline{2-14} & MTAN \cite{li2021multiattention} &  84.12 & 87.55 & 76.32 & 86.49 & 71.59 & 82.63 & 62.76 & 77.04 & 88.6  & 87.37 & 78.54 & 87.98\\
    \cline{2-14} & BAM \cite{park2018bam} & 84.27 &  87.83 & 76.45 & 86.55& 71.78 & 82.79 & 62.47 & 76.95 & 88.92 & 88.42 & 79.65 & 88.67\\
    \hline
    \multicolumn{2}{|c|}{\textbf{MSSDMPA-Net}} & \textbf{86.75} & \textbf{89.23} & \textbf{77.64} & \textbf{87.48}     &   \textbf{73.38}    &    \textbf{85.48}   &   \textbf{65.95}    & \textbf{79.50} & \textbf{91.11} & \textbf{90.11} & \textbf{82.83} & \textbf{90.61}\\
    \hline
    
    \end{tabular}%
    }
  \label{tab3}%
\end{table*}%

\begin{table*}[!htbp]
  \centering
  \caption{Performances of the Ablation models on building segmentation datasets}
    \vspace*{-3mm}
   \scalebox{0.7}{
    \begin{tabular}{|c|c|cccc|cccc|cccc|cccc|}
    \hline
    \multicolumn{2}{|c|}{\textbf{Dataset}} & \multicolumn{4}{c|}{\textbf{Massachusetts Building}} &  \multicolumn{4}{c|}{\textbf{Synthinel}} & \multicolumn{4}{c|}{\textbf{WHU Satellite I}} & \multicolumn{4}{c|}{\textbf{WHU Ariel Imagery}} \\
    \hline
    \multicolumn{2}{|c|}{\textbf{Methods}} & \textbf{Recall} & \textbf{Precision} & \textbf{IOU} & \textbf{F1 Score} & \textbf{Recall} & \textbf{Precision} & \textbf{IOU} & \textbf{F1 Score} & \textbf{Recall} & \textbf{Precision} & \textbf{IOU} & \textbf{F1 Score} & \textbf{Recall} & \textbf{Precision} & \textbf{IOU} & \textbf{F1 Score}\\
    \hline
    Without & LIP \cite{gao2019lip} & 89.94 &   90.10 & 81.93 & 90.25 & 91.90 & 91.03  & 84.08   & 91.63 & 80.47 & 77.11 & 64.89 & 78.85 &  94.57 & 95.13 & 91.64 & 95.68\\
    \cline{2-18} DAMIP &
    Maxpool & 89.57     &   89.73    & 81.67 &   89.89 & 91.67     &   90.76    & 83.77 &   91.16    &    80.11  &   76.63    & 64.42 & 78.42 & 94.72 & 94.97 & 90.37 & 94.88 \\
    \cline{2-18} & Avgpool & 88.82 & 89.11 & 80.07 & 89.01 & 91.63 & 90.61 & 83.61 & 91.09  & 80.05 & 76.59 & 63.84 & 78.09 & 94.79 & 94.67 & 89.98 & 94.72\\
    \cline{2-18} & Stochastic pool \cite{zeiler2013stochastic} & 88.67 & 88.98 & 80.05 & 88.99 & 91.55 & 90.40 & 83.48 & 90.93 & 79.94 & 75.41 & 63.67 & 77.91 & 94.43 & 94.82 & 89.74 & 94.65 \\
    \hline
    \multicolumn{2}{|c|}{Without Deep Supervision} & 87.14 &  86.74 & 76.82 & 86.83 & 90.39 & 89.25  & 81.57 & 89.93 & 79.71 & 75.86 & 63.56 & 77.75 & 93.19 & 93.51& 87.87  & 93.48 \\
    \hline
    \multicolumn{2}{|c|}{Without  Dilation} &    89.43   &   89.59    &   81.19    & 89.67  &    93.17   &   92.25    &   85.98    &    92.55   &   82.21    &    78.83   &    67.12   & 80.31 & 96.50 & 96.67 & 93.12 & 96.53 \\
    \hline
    \multicolumn{2}{|c|}{Without DAMSCA Block} & 88.79 & 88.68 & 79.97 & 88.71 & 91.89 & 90.95 & 84.25 & 91.38 & 81.25  & 76.71 & 65.27 & 78.90  & 94.25 & 94.41 & 89.10 & 94.32 \\
    \hline
    Different  & SAGAN \cite{zhang2019self}  & 90.40 & 90.25 & 82.67 & 90.45  & 92.78 & 91.79 & 85.78 & 92.34 &  82.14 & 77.76 & 66.57 & 79.92 & 95.59 & 95.87 & 91.89 & 95.76 \\
    \cline{2-18} self & CBAM \cite{woo2018cbam} & 90.06 & 90.11 & 81.82 & 90.10 & 92.57 & 91.33 & 84.90 & 91.86 & 81.95 & 77.53 & 65.99 & 79.61 & 95.13 & 95.41 & 91.23 & 95.31\\
    \cline{2-18} attention & MTAN \cite{li2021multiattention} & 89.81 & 89.85 & 81.44 & 89.82 &  92.20 & 90.97 & 84.27 & 91.44 & 81.54 & 77.07 & 65.89 & 79.33 & 94.42 & 94.76 & 89.89 & 94.73 \\
    \cline{2-18} pluggins & BAM \cite{park2018bam} & 89.87 & 89.93 & 81.78 & 89.94 & 92.36 & 91.16 & 84.54 & 91.62 & 81.67 & 77.31 & 66.08 & 79.45 & 94.79 & 94.92 & 90.01 & 94.84 \\
    \hline
    \multicolumn{2}{|c|}{\textbf{MSSDMPA-Net}} & \textbf{92.72} & \textbf{92.34} & \textbf{86.10} & \textbf{92.53} & \textbf{94.16} & \textbf{94.10} & \textbf{88.89} & \textbf{94.11} & \textbf{84.60} & \textbf{80.20} & \textbf{69.97} & \textbf{82.33} & \textbf{97.50} & \textbf{97.82} & \textbf{95.40} & \textbf{97.65} \\
    \hline
    \end{tabular}%
    }
  \label{tab4}%
\end{table*}%

Table \ref{tab1} presents the performances of the proposed and existing models on the road segmentation datasets Porto, Shanghai, and Massachusetts. Among all these existing models, the SOTA model, DeepDual Mapper depicts the mean IoU of 71.7\%, F1 score of 83.5\% on the Porto dataset and mean IoU of 63.4\%, F1 score of 77.6\% on the Shanghai dataset. The proposed model shows 77.64\% and 87.48\% of mean IoU and mean F1 score on the Porto dataset with an improvement of 5.94\% and 3.98\%, respectively, concerning the SOTA model DeepDual Mapper. The proposed model shows 65.95\% and 79.50\% of mean IoU and mean F1 score on the Shanghai dataset with an improvement of 2.55\% and 1.90\%, respectively, for the SOTA model DeepDual Mapper. The SOTA model of the Massachusetts Road dataset, ASSP Net shows an F1 score of 88.18\% and a mean IoU of 78.86\%. The proposed model shows a mean F1 score of 90.61\%, and a mean IoU of 82.83\% with an improvement of 2.43\% and 3.97\%, respectively.

\subsubsection{Results on building segmentation datasets}


Table \ref{tab2} depicts the performances of the MSSDMPA-Net and existing models on the building segmentation datasets. In Massachusetts dataset, the SOTA model BRRNet shows a mean IoU of 74.46\% and a mean F1 score of 85.36\%, and our MSSDMPA-Net shows 86.10\%, 92.53\% mean IoU and mean F1 score with an improvement of 11.64\%, 7.17\% respectively on this dataset. The MAPNet shows 82.03\%, 90.13\% of mean IoU and mean F1 score respectively on Synthinel dataset and the 63.05\%, 77.34\% of mean IoU and mean F1 score respectively on WHU Satellite I dataset. The proposed model shows 88.89\%, 94.11\% of mean IoU, mean F1 score with an improvement of 6.86\%, 3.98\% respectively on Synthinel dataset and 69.97\%, 82.33\% of mean IoU, mean F1 score with an improvement of 6.92\%, 4.99\% respectively on WHU Satellite I dataset. The Res2-Unet model shows 96.28\% and 92.83\% of the mean F1 Score and mean IoU, respectively, on WHU Ariel Imagery dataset. The proposed model outperforms the existing models with 97.65\% and 95.40\% of mean F1 Score and mean IoU with an improvement of 1.37\% and 2.57\%, respectively, to the SOTA model. \textit{The information-preserving nature of the MSSDMPA-Net along with its context awareness helped in generating feature maps with fine-grained structural details of the building polygons and road networks. This resulted in the MSSDMPA-Net to outperform all other state-of-the-art methods, along with the generation of stellar segmentation maps of roads and buildings.} 

%


\subsection{Ablation Studies} 
The Ablation studies depict the strength of different segments of deep learning models. To examine detailed performances, we have evaluated the model i) without DAMIP, ii) without deep supervision, iii) without dilation, and iv) without DAMSCA. The effectiveness of different modules of the proposed model is presented here.

\noindent\textbf{Effectiveness of DAMIP:} The effectiveness of the DAMIP module in preserving important features and reducing information loss has been examined by replacing the module with commonly used pooling mechanisms like LIP, max pool, average pool, and stochastic pool. 
Among all the pooling methods except for the proposed DAMIP, the LIP performs best, followed by Maxpooling, Average pooling, and Stochastic pooling in Porto, Shanghai, Massachusetts Building, Synthinel, WHU Satellite I and WHU Ariel Imagery datasets, and Maxpooling performs best in the Massachusetts road dataset.

\noindent\textbf{Effectiveness of DAMSCA:} To see the effectiveness of the DAMSCA module, we have removed the DAMSCA module, which shows the significance of the DAMSCA module. Additionally, we have substituted the self-attention mechanisms like SAGAN, CBAM, BAM, MTAN to compare with the DAMSCA module. 
Among the Attention blocks comparable to DAMSCA, the SAGAN performs best, followed by the CBAM, BAM, and MTAN. The proposed model outperforms all these models on these road and building datasets.

\noindent\textbf{Effectiveness of Deep supervision:} The efficacy of deep supervision has been analyzed by supervising the model only from the output. Thus, the effect of deep supervision stops at the DAMSCA and DAMIP modules. Thus, the effect of the removal of the deep supervision is very crucial for the model.
The proposed model shows improvement of 4.40\%, 5.57\%, 4.35\%, 5.70\% , 4.18\%, 4.58\%, 4.17\% in F1 score and 5.27\%, 7.33\%, 6.99\%, 9.28\%, 7.32\%, 6.41\%, 7.53\% in IoU on Porto, Shanghai, Massachusetts Road, Massachusetts Building, Synthinel-1, WHU Satellite I and WHU Ariel Imagery dataset over when the deep supervision is removed. The results show that deep supervision is the most effective component among all the novel components.

\noindent\textbf{Effectiveness of Dilation:} The effectiveness of the dilation is explored by examining the model with standard convolution operations in the encoders of each path of MSSDMPA-Net. 
The proposed model shows improvement of 1.56\%, 2.78\%, 1.79\%, 2.86\%, 1.56\%, 2.02\%, 1.12\% in F1 score and 2.48\%, 3.66\%, 2.94\%, 4.91\%, 2.91\%, 2.85\%, 2.28\% in IoU on Porto, Shanghai, Massachusetts Road, Massachusetts Building, Synthinel-1, WHU Satellite I and WHU Ariel Imagery dataset over when the dilation is removed.

\section{Conclusions}


In this paper, we propose a novel multi-path deep learning network, MSSDMPA-Net, which is equally effective in accurately segmenting road networks and building polygons from satellite images by leveraging the prior generated segmentation maps as attention masks. Towards this goal, we propose to reduce information loss during the spatial downsampling stages. This is achieved by making the downsampled multiscale features attentive to the semantic context of interest via the deployment of the novel DAMIP modules. Further, the novel DAMSCA modules upsample the spatial resolution of the dilated-convolution-refined multiscale feature maps to a higher spatial resolution which contains context-aware high-level semantic information. Finally, the multiscale features generated prior segmentation-maps are minutely supervised during training which not only helps in robust attention mechanism but also makes the MSSDMPA-Net immune to vanishing gradients. Our rigorous experimental evaluations on seven datasets convincingly confirm the efficacy of MSSDMPA-Net over the previous literature. The future direction may consider the notion of low-shot segmentation. 




 


{\footnotesize

\bibliographystyle{unsrtnat}

\bibliography{Main_Final.bib}}


 





\end{document}